\documentclass[journal]{IEEEtran}

\usepackage{algorithm}
\usepackage{algpseudocode}
\usepackage{multirow}
\usepackage{nicematrix}
\usepackage{graphicx}
\usepackage{cite}
\usepackage{tabu}

\usepackage[T1]{fontenc}
\usepackage{babel}
\usepackage{amssymb}
\usepackage{booktabs}
\usepackage{diagbox}
\usepackage{subfigure}
\usepackage{graphicx}
\usepackage{hyperref}
\DeclareRobustCommand\onedot{\futurelet\@let@token\@onedot}
\def\@onedot{\ifx\@let@token.\else.\null\fi\xspace}

\def\argmin{\operatornamewithlimits{arg\,min}}

\title{EVD4UAV: An Altitude-Sensitive Benchmark to Evade Vehicle Detection in UAV}

\author{Huiming Sun, 
Jiacheng Guo,
Zibo Meng,
Tianyun Zhang,
Jianwu Fang,
Yuewei Lin,
Hongkai Yu$^{*}$

  \thanks{ Huiming Sun, Jiacheng Guo, Tianyun Zhang and Hongkai Yu are with Cleveland State University, Cleveland, OH, 44115, USA. Zibo Meng are with OPPO US Research Center, Palo Alto, CA, 94303, USA. Jianwu Fang is with Xi'an Jiaotong University, Xi'an, 710049, China. Yuewei Lin is with Brookhaven National Laboratory, Upton, NY, 11973, USA.
  $^{*}$Corresponding author: Hongkai Yu  
(\href{mailto:h.yu19@csuohio.edu}{h.yu19@csuohio.edu})
}
  
  }

\begin{document}

\maketitle

\begin{abstract}
Vehicle detection in Unmanned Aerial Vehicle (UAV) captured images has wide applications in aerial photography and remote sensing. There are many public benchmark datasets proposed for the vehicle detection and tracking in UAV images. Recent studies show that adding an adversarial patch on objects can fool the well-trained deep neural networks based object detectors, posing security concerns to the downstream tasks. However, the current public UAV datasets might ignore the diverse altitudes, vehicle attributes, fine-grained instance-level annotation in mostly side view with blurred vehicle roof, so none of them is good to study the adversarial patch based vehicle detection attack problem. In this paper, we propose a new dataset named EVD4UAV as an altitude-sensitive benchmark to evade vehicle detection in UAV with 6,284 images and 90,886 fine-grained annotated  vehicles. The EVD4UAV dataset has diverse altitudes (50m, 70m, 90m), vehicle attributes (color, type), fine-grained annotation (horizontal and rotated bounding boxes, instance-level mask) in top view with clear vehicle roof. One white-box and two black-box patch based attack methods are implemented to attack three classic deep neural networks based object detectors on EVD4UAV. The experimental results show that these representative attack methods could not achieve the robust altitude-insensitive attack performance. 
The dataset could be found at  \href{https://github.com/HMS97/EVD4UAV}{\textcolor{red}{https://github.com/HMS97/EVD4UAV}}
\end{abstract}
    
\section{Introduction}
\label{sec:intro}

Unmanned Aerial Vehicles (UAVs) have been widely used in aerial photography~\cite{xie2023automating,sun2023coarse}, intelligent  transportation~\cite{yang2022classification, betz2022autonomous,kitajima2022nationwide,lin2022multi,emu2022fatality,xie2022safe}, and remote sensing~\cite{bansal2022rate, bharati2018real}. UAV captured images include rich visualized information on the road surfaces. Many public benchmark datasets are  proposed for the vehicle detection and tracking in UAV images, such as UAVDT~\cite{du2018unmanned}, VeRi~\cite{wang2019orientation,lyu2020uavid}, UAVid~\cite{lyu2020uavid} and UAVW~\cite{wang2022aprus}.

Based on the recent research of adversarial machine learning~\cite{thys2019fooling}, the deep neural networks based object detectors are found not robust to some adversarial  perturbations. Among them, the physical adversarial patch based detection attack arises severe security concerns to the downstream tasks. The physical adversarial patches can be printed~\cite{thys2019fooling} to put on the object to evade/fool the well-trained deep neural networks based object detectors.

\begin{figure}
\centering
\footnotesize
\includegraphics[width=1\columnwidth]{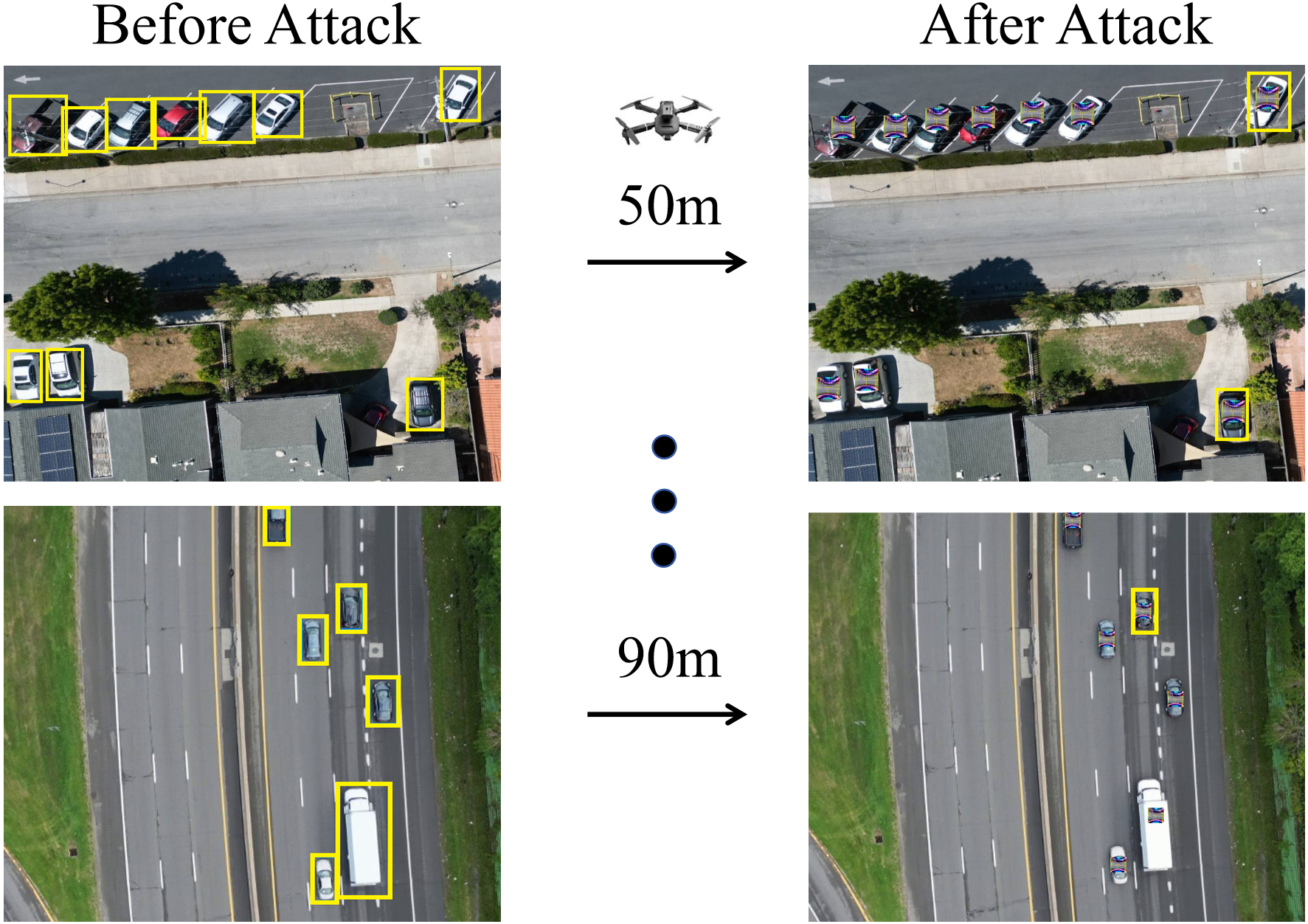}
\caption{Example vehicle detection before and after the adversarial attack in the altitudes  of 50m, 90m in the proposed altitude-sensitive EVD4UAV dataset. Due to the space limit, the altitude of 70m is skipped.}
\label{fig:motivation} 
\end{figure}

\begin{table*}[]
\centering
\caption{\textbf{Comparison of our EVD4UAV dataset and existing representative UAV datasets.} HBB: object annotation with Horizontal Bounding Boxes. RBB: object annotation with Rotated Bounding Boxes. iMask: object annotation with instance-level mask.}
\label{tab:dataset_compare}	
\resizebox{1 \textwidth}{!}{ 
\begin{tabular}{l|l|l|l|c|c|l|c|c|c|l|l}
\cline{1-12}
\toprule
\textbf{Dataset}        & \textbf{Purpose}   & \textbf{\begin{tabular}[c]{@{}l@{}} Images \\/ FPS\end{tabular}} & \textbf{Altitude (m)}  & \textbf{\begin{tabular}[c]{@{}l@{}}Altitude\\ Label\end{tabular}} & \textbf{\begin{tabular}[c]{@{}l@{}}Vehicle\\ Attributes\end{tabular}} & \textbf{\begin{tabular}[c]{@{}l@{}}Resolution\end{tabular}} & \textbf{\begin{tabular}[c]{@{}l@{}}HBB\end{tabular}} & \textbf{\begin{tabular}[c]{@{}l@{}}RBB\end{tabular}} & \textbf{\begin{tabular}[c]{@{}l@{}}iMask\end{tabular}} & \textbf{View} & \textbf{\begin{tabular}[c]{@{}l@{}}Snowy \\ landscape\end{tabular}} \\  \hline 
UAVDT~\cite{du2018unmanned}    & \begin{tabular}[c]{@{}l@{}}Detect \& Track \end{tabular}         & 80,000 / 30                                                          & Low/Medium/High                                                 & ×                                                                 & Type                                                                  & $1080 \times 540$                                                            & \checkmark                                                                              & ×                                                                           & ×                                                                     & Mostly side   & ×                                                               \\
VeRi~\cite{wang2019orientation}                    & \begin{tabular}[c]{@{}l@{}}Detect \& Track \end{tabular}           & 1,685 / 30                                                           & N/A                                                               & ×                                                                 & ×                                                                     & $2720 \times 1530$                                                           & \checkmark                                                                              & ×                                                                           & ×                                                                     & Mostly side   & ×                                                          \\
UAVid~\cite{lyu2020uavid}                  & \begin{tabular}[c]{@{}l@{}} Semantic  Segment\end{tabular}         & 300 / 0.2                                                            & 50                                                                & ×                                                                 & ×                                                                     & \begin{tabular}[c]{@{}l@{}}$4096 \times 2160$ \end{tabular}    & ×                                                                              & ×                                                                           & ×                                                                     & Mostly side   & ×                                                          \\
UAVW~\cite{wang2022aprus}                   & Detect                                                                & 46,037 / 30                                                          & \begin{tabular}[c]{@{}l@{}}50/60/70/80/90/100\end{tabular} & \checkmark                                                                 & ×                                                                     & $3840 \times 2160$                                                           & \checkmark                                                                              & ×                                                                           & ×                                                                     & Side          & ×                                                              \\

DroneVehicle~\cite{DBLP:journals/corr/abs-2003-02437}     &          Detect                            & 28439 / -                                                   & 80/100/120                                               & \checkmark                                                        &-                                                  & $840 \times 712$                                                  & \checkmark                                                                     &  \checkmark                                                                   & ×                                                           & Top  & ×      \\  
\hline
EVD4UAV & Detect Attack                                         & 10051 / 1                                                   & 50/70/90                                               & \checkmark                                                        & Color/Type                                                  & $1920 \times 1080$                                                  & \checkmark                                                                     & \checkmark                                                                  & \checkmark                                                            & Top  & \checkmark      \\  
\bottomrule
\end{tabular}
}
\end{table*}

However, all the existing public UAV datasets are not suitable to investigate the problem of evading vehicle detection in UAV images, because they are for vehicle detection and tracking~\cite{chen2023robust,sun2023vision, hickling2023robust, yu2024multi,masmoudi2021reinforcement} rather than patch-based detection attack. For the task of patch-based vehicle detection attack, altitude label is important because UAV flying height affects the object scale/size significantly, and keeping top-view images with clear vehicle roof is better for this study because the printed patch is more convenient to be put on the top of vehicles. The existing public UAV datasets might ignore the diverse altitudes, in mostly side view with the blurred vehicle roof. In addition, their annotations miss many details, vehicle attributes, and fine-grained instance-level annotation. 

In this paper, we propose a new dataset named EVD4UAV as an altitude-sensitive benchmark to \textbf{E}vade \textbf{V}ehicle \textbf{D}etection in \textbf{UAV} with 6,284 images and 90,886 fine-grained annotated vehicles. The comparison between the proposed EVD4UAV and existing representative UAV datasets is shown in Table~\ref{tab:dataset_compare}. Our EVD4UAV dataset has diverse altitudes (50m, 70m, 90m), vehicle attributes (color, type), fine-grained annotation (horizontal and rotated bounding boxes, instance-level mask) in top view with clear vehicle roof. One white-box and two black-box patch-based attack methods are implemented to attack three classic DNN based object detectors on EVD4UAV. As shown in Fig.~\ref{fig:motivation}, the vehicles wearing the learned adversarial patch cannot be detected by the classic object detectors, like YOLOv8~\cite{redmon2016you}, Faster R-CNN~\cite{ren2015faster}. Our extensive experimental results show that the representative attack methods could not achieve robust altitude-insensitive attack performance, so it is necessary to continue the research direction of altitude-insensitive attack using our EVD4UAV. The contribution of this paper can be summarized as follows.

\begin{itemize}
    \item A new dataset with fine-grained annotations named EVD4UAV is proposed as an altitude-sensitive benchmark to evade vehicle detection in UAV. 

    \item Using the proposed EVD4UAV, this paper investigates representative attacks (one white-box and two black-box methods) to test the detection evasion performance in UAV. 

    \item Using the proposed EVD4UAV for extensive experiments, this paper points out that the altitude-insensitive detection attack in UAV is worth further investigation. 
\end{itemize}

\section{Related Work} 

\subsection{UAV Object Detection}

Aiming at improving clarity with reasonable accuracy and higher surveying efficiency, Unmanned Aerial Vehicles (UAV) has become an advanced way to deal with object detection tasks. UAVs can be categorized into two classes: fixed-wing UAVs and rotary-wing UAVs~\cite{zuo2022unmanned}. A typical usage of UAV is UAV-aided information dissemination and data collection by wireless communication technologies~\cite{zeng2016wireless}. Those UAVs have sustainable speed when flying across large certain areas in a short time, which could collect and process data in high efficiency~\cite{zeng2016wireless}~\cite{zuo2022unmanned}. However, it is pointed out that UAV object detection may suffer from hostile natural environments, camera issues, and illumination issue challenges in actual application operation~\cite{shakhatreh2019unmanned}  \cite{xu2023v2v4real,sun2021convolutional,sun2023coarse,sun2024defense,liu2023deep,liu2023deep,yin2023air}. In the last decade, many researchers have provided insights into these challenges. In the first stage, object detection is correlated with Deep Neural Networks (DNNs) in \cite{szegedy2013deep}, which obtains high-resolution object detection images at a low cost. Later, a benchmark was established in~\cite{redmon2016you}, where the proposed “You Only Look Once” (YOLO) structure is sustainable for large-scale single-shot object detectors. The bounding box, class score and object score could be directly predicted by only a single pass going through the structure~\cite{yolov8_ultralytics}~\cite{redmon2016you}. Moreover, an end-to-end object relation detector is built in~\cite{hu2018relation} which further improves the detecting accuracy by duplicating removal steps. Recently, different vehicle detection frameworks under challenging scenarios has been proposed~\cite{li2023domain,yu2024sora,liu2023deep}.

Overall, the accuracy of object detection has been improving in recent years using DNN models. However, most existing UAV image datasets are designed for object detection and constructing datasets for adversarial attacks remains rare. This paper aims to propose a new benchmark to attack the object detectors using images taken by UAV. 

\subsection{Adversarial Attack in Object Detection}
Adversarial attacks could be divided into two categories: digital attack and physical attack~\cite{hu2021naturalistic}. Digital attack means the attack input is generated by digital attack images; physical attack means the attack images which are taken as the attack input are retaken by a camera~\cite{hu2021naturalistic}. The first application is proposed by \cite{xie2017adversarial} named Dense Adversary Generation (DAG) algorithm which is based on optimization methods. DAG established adversarial labels in each proposal region and then misclassified the proposals using iterative gradient back-propagation. However, DAG is time-consuming and needs many iterations to go through an adversarial image. Later, the adversarial attack was applied on face detectors in~\cite{bose2018adversarial}, but the model still relies on proposal-based detectors. Besides adding noise among the whole image, a number of methods only modify part of regions of the image to attack object detectors~\cite{li2018exploring}~\cite{liu2018dpatch}. However, most of these methods give little improvement. A more enhanced generated model is proposed by \cite{zhao2019seeing} whose detectors with adversarial examples could be used in multiple scenes in the real world with higher robustness and transferability. The first adversarial patches generated for attacking intra-class variety was established in \cite{thys2019fooling}, taking persons as the specific class. By generating patches, they propose adversarial attacks with much higher accuracy. Moreover,~\cite{hu2021naturalistic} generated adversarial patches with are more naturalistic looking and less doubtful. While ensuring higher attack accuracy, the adversarial patches are more hidden and less noticeable.

In general, the current public UAV datasets may ignore the diverse altitudes, vehicle attributes, and fine-grained instance-level annotation. However, these datasets might not be effectively used for adversarial attack tasks with mostly side view and blurred vehicle roofs. Therefore, none of them is good enough to study the adversarial patch-based vehicle detection attack problem. In this paper, we propose a new dataset named EVD4UAV as an altitude-sensitive benchmark with top view and clear roof to evade vehicle detection in UAV.

\section{Benchmark}

\subsection{Data Collection}

All the images in our EVD4UAV dataset are taken from UAV cameras. The images are all from a top-view overhead perspective taken from three different altitudes: 50m, 70m, and 90m above the ground. Each image contains several moving and stable vehicles and each vehicle is labeled with fine-grained annotation. The UAV images were taken in California and Ohio, USA. We use DJI Mini3 Pro to capture sparsely sampled videos (FPS=1) and some discrete single images with a camera degree of nearly vertical view angle. We focus on the object detection task and categorize the images into four different scenes based on the location of vehicles: Urban Road, Highway, Parking Lot, and  Residential Area, reflecting the diversity of vehicle activities. 

\begin{figure}[htbp]
\centering
\footnotesize
\includegraphics[width=1\columnwidth]{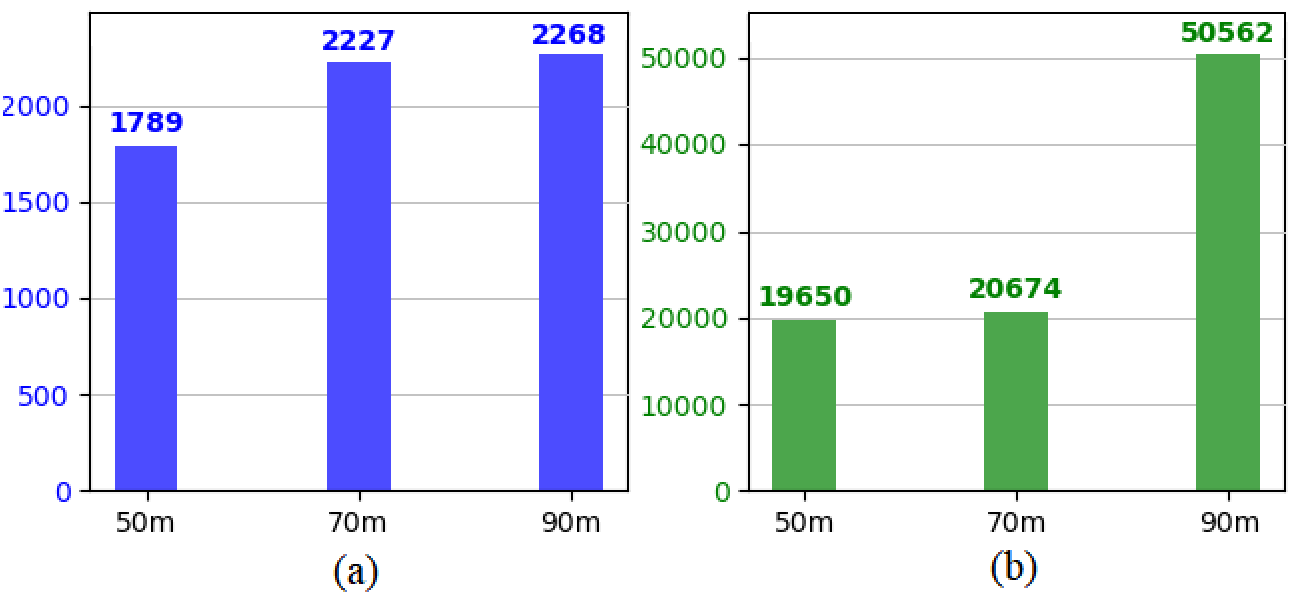}
\caption{Total number of (a) images and (b) fine-grained annotated vehicles of our EVD4UAV in different UAV altitudes (50m, 70m, 90m).}
\label{fig:imgs and bounding boxes}	 
\end{figure}

Specifically, we have a total of 6,284 images of \(1,920 \times 1,080 \) resolution with a total number of 90,886 fine-grained annotated vehicles. The total number of images and annotated vehicles from different altitudes is shown in Fig.~\ref{fig:imgs and bounding boxes}. 1,789 images are taken from altitude 50m above the ground, and 2,227 and 2,268 images are taken from 70m and 90m respectively. The number of fine-grained annotated vehicles of 50m and 70m are around 20,000 each, and the images taken from altitude 90m has more than 50,000 due to a larger perception field. 

\begin{figure*}
\centering
\footnotesize
\includegraphics[width=1\textwidth]{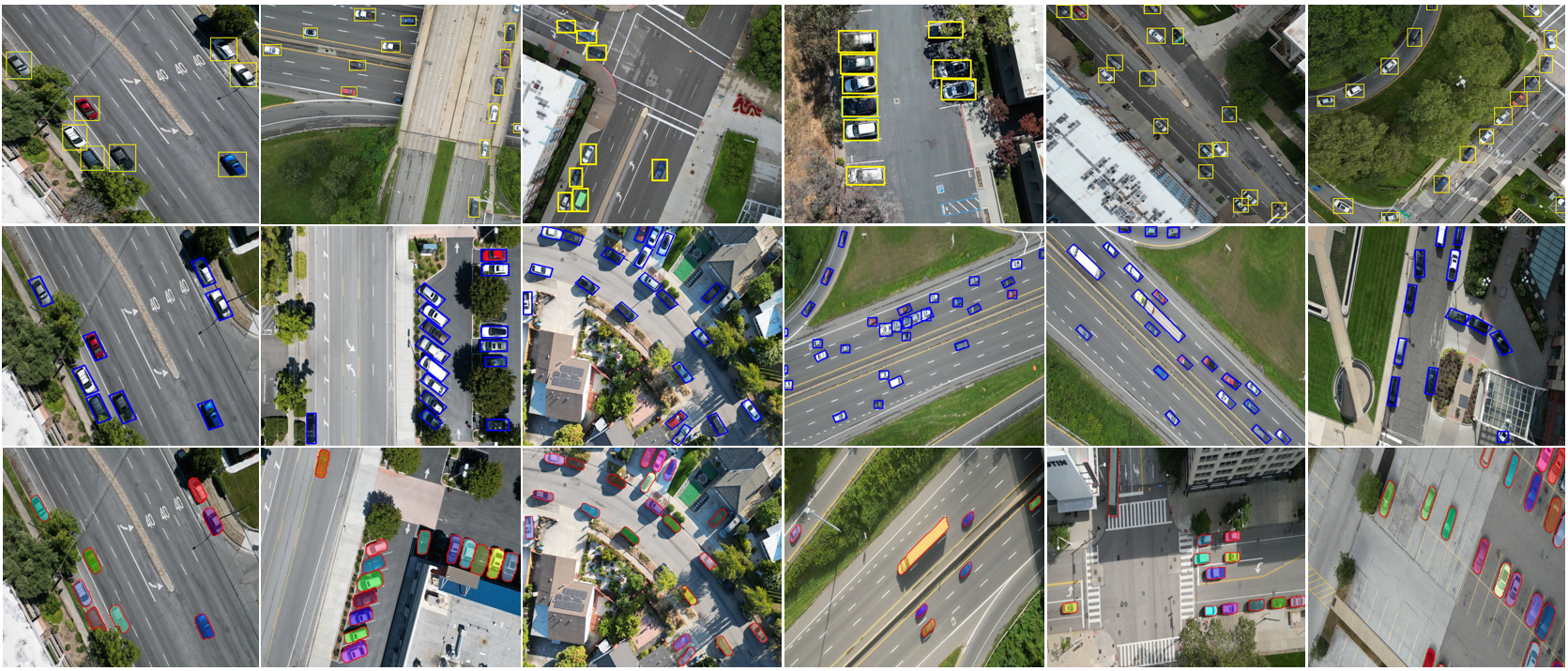}
\caption{Example vehicle images in EVD4UAV with labeled horizontal bounding boxes, rotated bounding boxes, and instance-level segmentation masks from top to bottom.}
\label{fig:Images with bounding boxes and masks.}	 
\end{figure*}

\begin{figure*}
\centering
\footnotesize
\includegraphics[width=\textwidth]{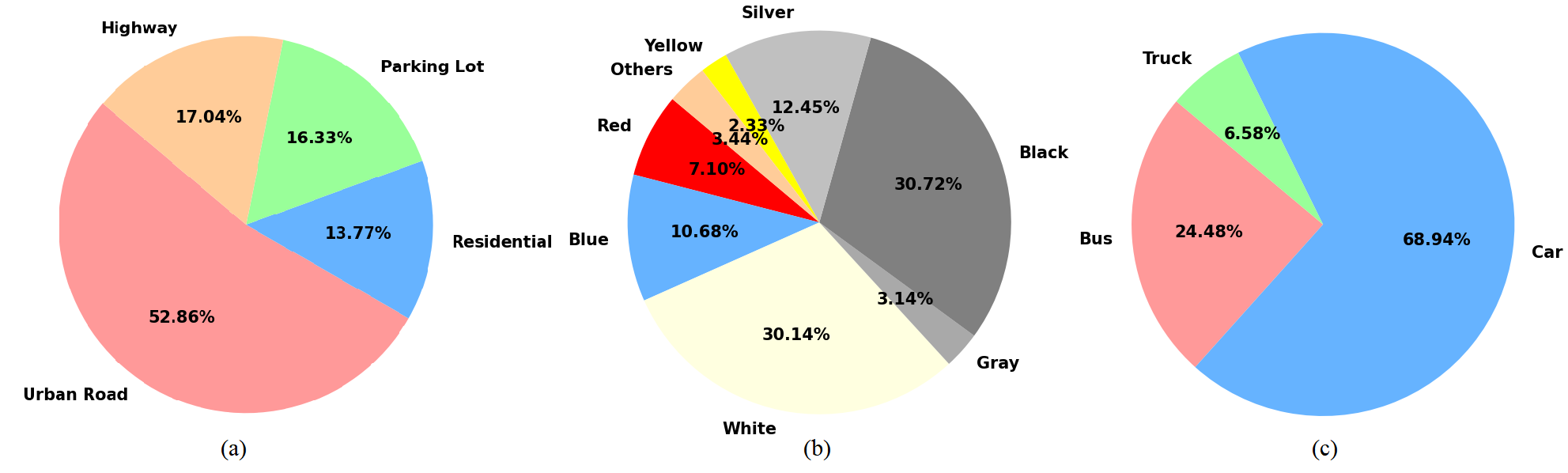}
\caption{Proportions of (a) images in four scenes, (b) annotated vehicle color attribute, (c) annotated vehicle type attribute in EVD4UAV.}
\label{fig:pie charts.}	 
\end{figure*}

We labeled horizontal and rotated bounding boxes for each vehicle in the images in the proposed EVD4UAV. Besides, we accomplish an instance-level segmentation task for each vehicle as shown in Fig.~\ref{fig:Images with bounding boxes and masks.}. The first row shows the vehicles labeled with horizontal bounding boxes. The second and third row shows the vehicles labeled with rotated bounding boxes and instance-level segmentation masks respectively. Vehicles under various scenes can be seen in different columns.

\subsection{Data Analysis}


We categorized the images into four scenes based on where the vehicles are: on the Urban Roads, on the Highways, in the Parking Lots, and in the Residential Areas, with the proportion of each scene shown in Fig.~\ref{fig:pie charts.}(a). About half of the vehicles are mainly on urban roads. The proportions of the remaining three categories are relatively even, at around 13\% to 17\%.

Our EVD4UAV dataset also has a diversity of vehicles under various weather conditions and brightness. The proportions of images taken from various weather conditions. Most of the images are taken in cloudy weather, mostly cloudy days in Ohio. One-fifth of the images are taken on sunny days and about  10\% of the images are taken at dusk. Leveraging images under different weather conditions and brightness conditions helps build more robust adversarial attack models in various situations.

The Fig.~\ref{fig:pie charts.}(b) and Fig.~\ref{fig:pie charts.}(c) illustrate the annotated vehicle attributes in colors and types in our EVD4UAV dataset respectively. As shown in Fig.~\ref{fig:pie charts.}(b), there are seven common colors of vehicles, and the largest two proportions are Black and White colors. As the types illustrated in Fig.~\ref{fig:pie charts.}(c), most of the vehicles contained in the images belong to Cars, with nearly a quarter of the vehicles belonging to Buses and a small percentage of Trucks.


\begin{figure*}[htbp]
\centering
\footnotesize
\includegraphics[width=0.95\textwidth]{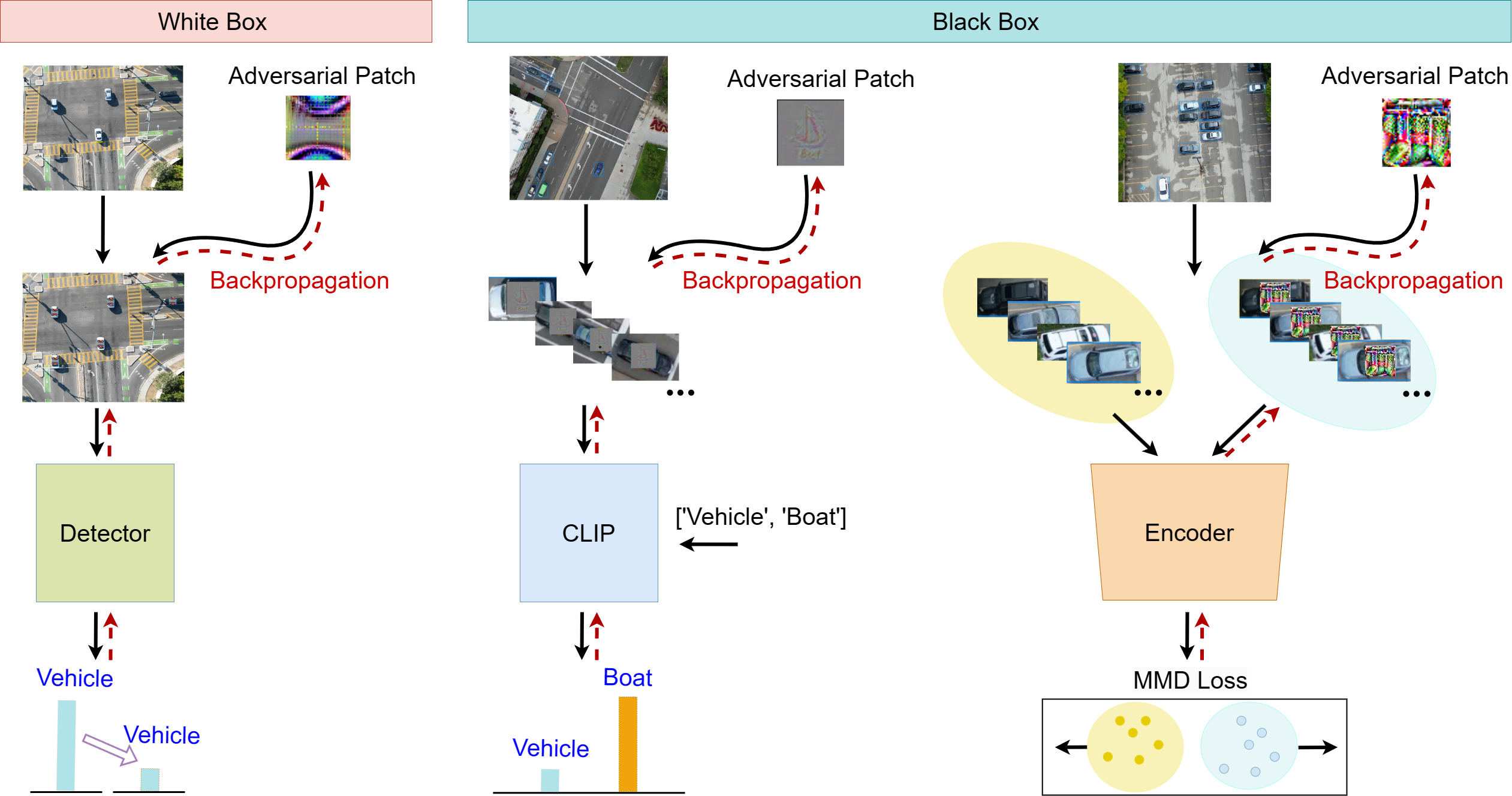}
\caption{Schematic representation of adversarial patch generation and application processes for white-box and black-box attacks on UAV vehicle detectors, highlighting the path from patch creation to detector evasion and classification discrepancy.}
\label{fig:pipeline}	 
\end{figure*}

\section{Methodology}
This paper aims to assess the effectiveness of adversarial patches in evading vehicle detection from various altitudes using UAVs. We focus on both white-box and black-box attack scenarios to understand the robustness of current UAV detection models under adversarial conditions. The pipeline of our designed representative white-box and black-box attack methods are shown in Fig.~\ref{fig:pipeline}. 


\subsection{Adversarial Patch Generation}
The adversarial patch is initialized as a tensor $P$  with the dimension $ \mathbb{R}^{H \times W \times 3}$. \textit{Our optimization goal is to learn a unified adversarial patch adding to any vehicle tops so as to evade the vehicle detection in UAV, one learned  adversarial patch for the whole dataset}. We constrain that the patch size is less than 40\% of the mean value of the Gaussian distributions of the vehicle areas in corresponding altitudes. During training, the unified patch is added to the center of each vehicle top on the original UAV image, and then the manipulated UAV image is sent to the object detection model. The patch $P$ is then learned through a training procedure using Stochastic Gradient Descent (SGD) as the optimizer. The patch updating process during optimization is defined as:
\begin{equation}
    P^{i+1} = P^{i} - \eta \cdot \nabla_{P} \mathcal{L}_a, 
\end{equation}
where $P^{i} $ is the learned adversarial patch with the batch identity $i$, $\eta$ denotes the learning rate, and $\nabla_{P}$ denotes the gradient of the corresponding attack loss function $\mathcal{L}_a$ with respect to the $P$. The details of the attack loss functions $\mathcal{L}_a$ in different settings will be described in the following.

\subsection{White-Box Patch (WBP) Attack}
This part focuses on developing white-box patch based attacks targeting at classic object detectors in UAV images. Similar to~\cite{thys2019fooling}, the goal of optimizing the adversarial patch is to minimize the inverse of the corresponding object class loss by directly feeding the original UAV image plus the added patches on each vehicle into the corresponding detectors. In this paper, we studied three classic object detectors, Faster R-CNN~\cite{ren2015faster}, DETR~\cite{carion2020end},YOLOv8~\cite{redmon2016you}. Because of their different architectures, we define different attack loss functions $\mathcal{L}_a$  respectively.

\textbf{$\text{WBP}_\text{RCNN}$:} If the target network is Faster R-CNN~\cite{ren2015faster}, we define the attack loss function as:  
\begin{equation}
    \mathcal{L}_a= \frac{1}{N} \sum_{i=1}^{N} s_i,  
\end{equation}
where $s_i$ is the score/confidence of the $i$-th region proposal, and $N$ represents the number of region proposals generated by the Region Proposal Network (RPN) in Faster R-CNN prediction. Minimizing this loss function will reduce the score/confidence of all the generated object proposals by RPN in Faster R-CNN.

\textbf{$\text{WBP}_\text{DETR}$:} If the target network is DETR~\cite{carion2020end}, we define the attack loss function as:
\begin{equation}
\mathcal{L}_a =  -1 \cdot 
 \sum_{i=1}^{m}\sum_{j=1}^{n} - y_{j} \log(s_{ij}),  
\end{equation}
where the right part is the normal cross-entropy loss for the correct classification of the predicted bounding boxes of multiple feature stages, $m$ represents the number of feature stages, $n$ represents the number of predicted bounding boxes, $y$ represents the object class label, $s$ represents the predicted score/confidence. Minimizing this loss function will confuse the correct classification of the predicted bounding boxes in DETR.

\textbf{$\text{WBP}_\text{YOLOv8}$:} If the target network is YOLOv8~\cite{redmon2016you}, we define the attack loss function as:
\begin{equation}
\mathcal{L}_a = -1 \cdot \frac{ -\sum_{i=1}^{n} \left[y_{i} \cdot \log(s_{i}) + (1 - y_{i}) \cdot \log(1 - s_{i})\right]}{N_t}
\end{equation}
where the right part is the normal binary cross-entropy loss for the correct classification of predicted bounding boxes, $n$ represents the number of predicted bounding boxes, $y$ represents the object class label, $s$ represents the predicted object score/confidence, and $N_t$ represents the normalization term defined in YOLOv8. Minimizing this loss function will improve the misclassification of the predicted bounding boxes in YOLOv8.

    





In summary, the optimization procedure of WBP methods to discover the adversarial patch is denoted as below: 
\begin{align}\label{eq:adv_obj}
\argmin_{P}\mathcal{L}_a(\delta(\phi(I_{clean},P)), y), 
\end{align}
where $P$ represents the adversarial patch to be learned,   $\mathcal{L}_{a} $ represents the attack loss function,  $\delta$ represents the targeting object detector, $\phi$ represents the patch adding process, $I_{clean}$ represents the original clean  image, and $y$ represents the detection ground truth. Our $\text{WBP}$ method requires the targeting object detection model to compute gradients for optimization, so it is a white-box attack. 

\begin{figure*}[th]
\centering
\includegraphics[width=1\textwidth]{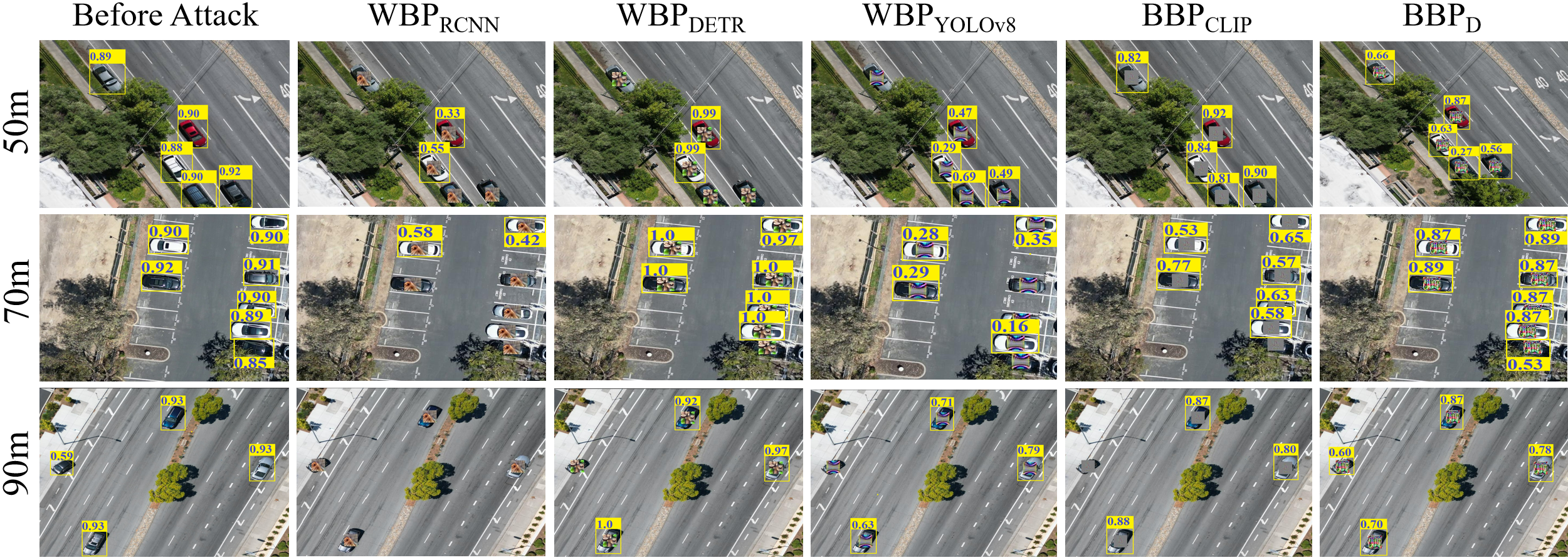}
\caption{ The detect bounding boxes and confidence results predicted by different models.  The $\text{BBP}_\text{CLIP}$ and $\text{BBP}_\text{D}$ are detected by YOLOv8 detector. Other attack methods are targeting their own detectors. }
\label{fig:imgs_models}	 
\end{figure*}

\subsection{Black-Box Patch (BBP) Attack}
In this section, we introduce two BBP attack methods,  Language-Image Model based BBP called \textbf{$\text{BBP}_\text{CLIP}$} and Distribution based BBP called \textbf{$\text{BBP}_\text{D}$}.


\subsubsection{Language-Image Model based BBP}

We employ a Language-Image contrastive learning framework,  the state-of-the-art CLIP model~\cite{radford2021learning}. CLIP (Contrastive Language-Image Pretraining) utilizes a multimodal approach to understand images in the context of natural language descriptions. In our framework, CLIP's visual image encoder, $ \text{CLIP}_{\text{VE}} $, processes the cropped vehicle image Region of Interests (RoIs), while its text encoder, $ \text{CLIP}_{\text{TE}} $, encodes the textual descriptions corresponding to object classes. Specifically, the cropped vehicle image RoIs added with the adversarial patches, denoted as $\mathbf{R_{adv}}$, are passed through the CLIP image encoder. Let us define a text [`vehicle', `confusion class'] as $\mathbf{T}$, then we use the CLIP encoders to extract the multimodal features:      
\begin{equation}
    F_{img}^{adv} = \text{CLIP}_{\text{VE}}(\mathbf{R_{adv}}),
\end{equation}
\begin{equation}
    F_{text} = \text{CLIP}_{\text{TE}}(\mathbf{T}).
\end{equation} 

\textbf{Score/confidence Calculation:} The score/confidence of each vehicle image RoI with the adversarial patch belonging to the respective text classes in $\mathbf{T}$ can be calculated via contrastive learning as:  
\begin{equation}
    s_{\mathbf{T}} = \mathbf{softmax}(\frac{F_{img}^{adv}}{\| F_{img}^{adv} \|_2} \cdot \frac{F_{text}^{T}}{\| F_{text} \|_2}). 
\end{equation} 

\textbf{Attack Loss Function:} For the attack purpose, our optimization aims to reduce the score/confidence of the `vehicle' class (making it close to 0) and increase the score of the `confusion class' class (making it close to 1). This goal could be achieved by minimizing the following Mean Square Error (MSE) loss: 
\begin{align}
    \mathcal{L}_a = \frac{1}{K} \sum_{k=1}^{K} (s_{\mathbf{T}} - [0,1])^2,   
\end{align}
where $K$ is the number of total cropped vehicle image RoIs in the training set. The optimization process is denoted: 
\begin{align}
    \argmin_{P}\mathcal{L}_a(s_{\mathbf{T}}).  
\end{align}

Minimizing this loss function serves to reduce the likelihood of correct class (`vehicle') and increase its probability to be the `confusion class'. In our experiment, we use `boat' as the `confusion class'.

\begin{table}[h]
\centering
\caption{The mAP50 results (before attack) for different Object Detection models on EVD4UAV dataset.}
\begin{tabular}{lcccc}
\toprule
\textbf{Model} & 50m & 70m & 90m  & Mean \\ \hline
Faster R-CNN~\cite{ren2015faster} & 0.9503 & 0.9473 & 0.9573 & 0.9552 \\ 
DETR~\cite{carion2020end} & 0.9524 & 0.9314 & 0.9145&    0.9263 \\ 
YOLOv8~\cite{redmon2016you} & \textbf{0.9824} & \textbf{0.9893} & \textbf{0.9912}& \textbf{0.9880} \\ 
\bottomrule
\end{tabular}
\label{tab:detection_result}
\end{table}

\begin{table*}[!]
\centering
\footnotesize
\caption{Detection results (mAP50) of EVD4UAV across various altitudes, using different adversarial patches by different detectors.}
\begin{tabular}{lcccc|cccc|cccc}
\toprule
\textbf{Attack} & \multicolumn{4}{c|}{Faster R-CNN~\cite{ren2015faster}}   & \multicolumn{4}{c|}{DETR~\cite{carion2020end}}  & \multicolumn{4}{c}{YOLOv8~\cite{redmon2016you}}  \\ \cline{2-13} 
\textbf{Methods}  & 50m        & 70m       & 90m &Mean      & 50m        & 70m       & 90m   &Mean    & 50m        & 70m       & 90m  &Mean     \\  \hline
Naive Cover & 0.9119 & 0.9132 & 0.9044 & 0.9098 & 0.9443 & 0.9226 & 0.9098 & 0.9256 & 0.9700 & 0.9812 & 0.9878 & 0.9797 \\ 

Random Noise & 0.9296 & 0.9366 & 0.9374 & 0.9345 & 0.9382 & 0.9169 & 0.9007 & 0.9186 & 0.9526 & 0.9578 & 0.9799 & 0.9634 \\ 
$\text{WBP}_\text{RCNN}$ & \textbf{0.1746} & \textbf{0.2895} & \textbf{0.5110} & \textbf{0.3250} & 0.8873 & 0.8541 & 0.8662 & 0.8692 & 0.9557 & 0.9422 & 0.9759 & 0.9579 \\ 
$\text{WBP}_\text{DETR}$ & 0.7368 & 0.8560 & 0.8638 & 0.8189 & 0.5069 & 0.7394 & 0.8749 & 0.7071 & 0.9174 & 0.8844 & 0.9705 & 0.9241 \\ 
$\text{WBP}_\text{YOLOv8}$ & 0.5114 & 0.6846 & 0.9239 & 0.7066 & \textbf{0.3127} & \textbf{0.5386} & \textbf{0.8623} & \textbf{0.5712} & \textbf{0.3367} & \textbf{0.1004} & \textbf{0.7453} & \textbf{0.3941} \\ 
$\text{BBP}_\text{CLIP}$ & 0.9296 & 0.9135 & 0.9164 & 0.9198 & 0.9111 & 0.8112 & 0.8820 & 0.8681 & 0.9401 & 0.8300 & 0.9718 & 0.9139 \\ 
$\text{BBP}_\text{D}$ & 0.9289 & 0.9458 & 0.9473 & 0.9407 & 0.9111 & 0.9349 & 0.8966 & 0.9142 & 0.9539 & 0.9643 & 0.9796 & 0.9659 \\ 
\bottomrule

\end{tabular}
\label{tab:results}
\end{table*}

\begin{table*}[h]
\centering
\footnotesize
\caption{Comparison of object detection performance (mAP50)  with different patch sizes with $\text{WBP}_\text{YOLOv8}$ attack.}
\begin{tabular}{lcccc|cccc|cccc}
\toprule
\textbf{Patch}  & \multicolumn{4}{c|}{Faster R-CNN~\cite{ren2015faster}} & \multicolumn{4}{c|}{DETR~\cite{carion2020end}} & \multicolumn{4}{c}{YOLOv8~\cite{redmon2016you}} \\ \cline{2-13} 

 \textbf{Size (pixel)}          & 50m    & 70m    & 90m    & Mean     & 50m    & 70m    & 90m    & Mean   & 50m    & 70m    & 90m    & Mean   \\ \hline
50/40/30   & \textbf{0.5114} & \textbf{0.6846} & \textbf{0.9239} & \textbf{0.7066}   & \textbf{0.3127} & \textbf{0.5386} & \textbf{0.8623} & \textbf{0.5712} & \textbf{0.3367} & \textbf{0.1004} & \textbf{0.7453} & \textbf{0.3941} \\ 
40/30/20   & 0.8486 & 0.9426 & 0.9481 & 0.9131   & 0.9349 & 0.9133 & 0.8966 & 0.9149 & 0.3466 & 0.3994 & 0.9856 & 0.5772 \\ 
30/20/10   & 0.9409 & 0.9472 & 0.9278 & 0.9386   & 0.9446 & 0.9290 & 0.9129 & 0.9288 & 0.8420 & 0.9900 & 0.9842 & 0.9387 \\ \bottomrule
\end{tabular}
\label{tab:ablationstudy}
\end{table*}

\subsubsection{Distribution based BBP}
In this section, we detail our approach of a feature distribution-based adversarial attack targeting UAV-based object detection systems.

\textbf{Feature Extraction:} The cropped vehicle image RoIs are fed into a pre-trained encoder model $\mathbf{E}$ for feature extraction. The encoder $\mathbf{E}$ is a neural network pre-trained on a large dataset and is capable of extracting complex features from images. We use the ResNet50 model pre-trained on ImageNet as $\mathbf{E}$ in our experiment. The features from both attacked and clean RoIs are extracted separately via:
    \begin{equation}
    \begin{split}
        \mathbf{F_{adv}} & = \mathbf{E}(\mathbf{R_{adv}}),  \\
        \mathbf{F_{clean}} & = \mathbf{E}(\mathbf{R_{clean}}),
        \end{split}
    \end{equation}
where $\mathbf{F_{adv}}$ and $\mathbf{F_{clean}}$ represent the features extracted from attacked vehicle image RoI set $\mathbf{R_{adv}}$ and the original clean  vehicle image RoI set $\mathbf{R_{clean}}$, respectively.  

\textbf{Feature Distribution Discrepancy:} We use the Maximum Mean Discrepancy (MMD)~\cite{gretton2006kernel} metric to quantify the difference in the feature distributions between the attacked and clean vehicle image RoIs. The MMD metric is particularly effective in measuring the distance between high-dimensional feature distributions~\cite{gretton2006kernel}. A higher MMD value indicates a greater divergence between the attacked and clean image RoIs, suggesting a more effective adversarial attack, which is computed as:
\begin{equation}
    \mathcal{L}_a = -1 \cdot \mathbf{MMD}(\mathbf{F_{adv}}, \mathbf{F_{clean}}), 
\end{equation}
where $\mathbf{MMD}(\cdot)$ denotes the function computing the MMD metric between two feature distributions. Then the attack optimization is defined as:
\begin{align}
    \argmin_{P}\mathcal{L}_a(\mathbf{F_{adv}}, \mathbf{F_{clean}}),   
\end{align}
whose goal is to enlarge the feature distribution difference between the cropped vehicle image RoIs added with the adversarial patch and the cropped original vehicle image RoIs.

In summary, our $\text{BBP}_\text{CLIP}$ and  $\text{BBP}_\text{D}$ methods do not require object detection models, so they are black-box attacks.

\section{Experiment}

\subsection{Object Detection Result}

The experimental results of the Object Detection are summarized in Table~\ref{tab:detection_result}. Faster R-CNN~\cite{ren2015faster}, DETR~\cite{carion2020end}, and YOLOv8~\cite{redmon2016you} are trained to evaluate the performance at different altitudes (50m, 70m, 90m). We keep all parameters default and set total epochs as 100. The training set and testing are set to 1:1 in EVD4UAV dataset. We treat all classes as one class ``Car''.

\subsection{Implementation Details}

Each of the input images has a resolution of $1920 \times 1080$. For altitudes of 50m, 70m, and 90m, the size of $P$ is set to $50 \times 50$, $40 \times 40$, and $30 \times 30$ pixels respectively. We initialize $P$ with the value of 0.5 in size of $50 \times 50$. We use it as the starting point of backpropagation to update, then directly resize it to $40 \times 40$ and $30 \times 30$ at the corresponding altitudes. For the training process, we employed the SGD optimizer. The training was configured with specific hyperparameters: a batch size of 4 and 100 epochs. Different learning rates were applied for various attack methods: 0.1 for  $\text{WBP}_\text{YOLOv8}$ $\text{BBP}_\text{CLIP}$, and $\text{BBP}_\text{D}$, and 0.4 for $\text{WBP}_\text{RCNN}$ and $\text{WBP}_\text{DETR}$. We also used a learning rate scheduler which reduces the rate by $50\%$ for every 20 epochs to optimize the performance. In the context of white box attack patch training, we resized the images to match the default size of each pre-trained detector to calculate the corresponding loss. All experiments were using 8 NVIDIA RTX A6000 GPU cards.

\subsection{Experiment Evaluation}

\begin{figure}
\centering
\footnotesize
\includegraphics[width=1\columnwidth]{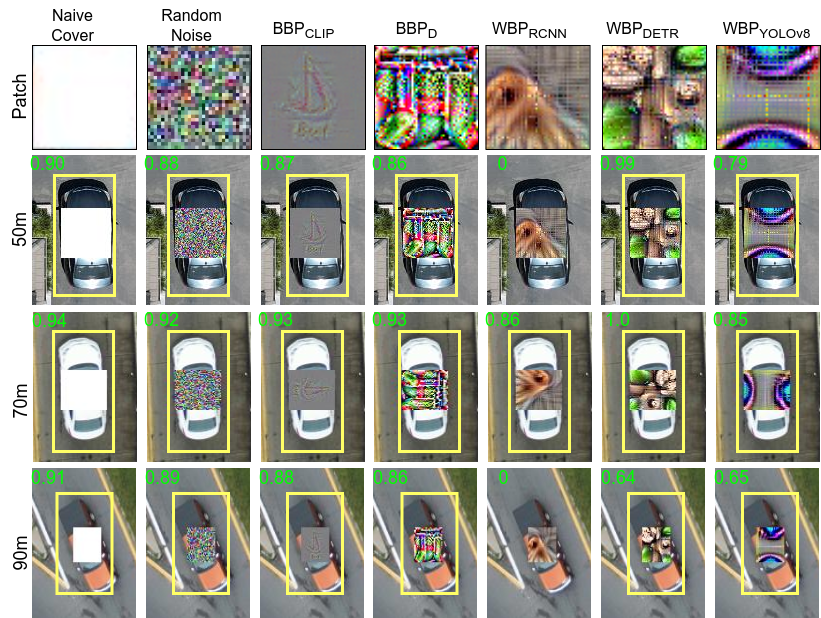}
\caption{Example images of attack patches and confidence score on different altitudes. The detection results of $\text{BBP}_\text{CLIP}$, $\text{BBP}_\text{D}$, Naiver Cover, and Random Noise are detected by YOLOv8 detector. Other attack methods are targeting their own detectors.  }
\label{fig:patches}	
\vspace{-15pt}
\end{figure}

The experiments conducted for the EVD4UAV dataset offer insights into the robustness of object detection models against adversarial attacks at varying altitudes. Table~\ref{tab:results} presents the mean Average Precision (mAP50) scores for different models when exposed to various adversarial patches, including those attack methods $\text{WBP}_\text{RCNN}$, $\text{WBP}_\text{DETR}$, $\text{WBP}_\text{YOLOv8}$, $\text{BBP}_\text{CLIP}$, $\text{BBP}_\text{D}$, Naive Cover, and Random Noise across altitudes of 50m, 70m, and 90m. The visualization on different $P$ examples is shown in Fig.~\ref{fig:patches}. As Fig.~\ref{fig:patches} shows, both black box attacks and white box attacks can serve to reduce performance, but the effect of the white box is far superior to the black box. \textit{Besides, there is no existing method that could achieve a good enough attack ability across three altitudes, so it is necessary to conduct more in-depth future research of the altitude-insensitive attack to UAV detection.}

The results are indicative of the models' susceptibility to $P$ shown in Fig.~\ref{fig:imgs_models}, with a noticeable decline in mAP50 scores as adversarial patches are applied. $\text{WBP}_\text{RCNN}$, when attacked itself, exhibits a significant increase in mAP50 from 0.1746 to 0.511 as the altitude changes from 50m to 90m, suggesting that altitude plays a role in the effectiveness of the adversarial patches. However, when exposed to adversarial patches generated by $\text{WBP}_\text{YOLOv8}$, its performance drops markedly at lower altitudes (0.5114 at 50m), but less so at higher altitudes (0.9239 at 90m).

\subsection{Ablation Study}

To assess the influence of patch size on the performance of object detection algorithms at different altitudes, we conducted an extensive ablation study. The experiments were performed with adversarial patches of different sizes ($50 \times 50$, $40 \times 40$, $30 \times 30$ pixels, decreasing the size of the patch with each experiment). We directly use the $P$ obtained from the YOLOv8 attack. Table~\ref{tab:ablationstudy} presents the mAP50 detection results. The performance of all models generally decreases with smaller patch sizes, illustrating the challenge of using smaller adversarial patches, especially at higher altitudes. Notably, Faster R-CNN showed a significant drop in attack performance at 90m when the patch size was reduced to $30 \times 30$, indicating sensitivity of patch size to higher altitudes.

\section{Conclusion}
This study presents the EVD4UAV dataset, a novel approach for evading vehicle detection in UAV imagery, highlighting the critical influence of altitude on the effectiveness of adversarial attacks. The experiments demonstrate that current object detection models are susceptible to such attacks, particularly at varying altitudes. This research not only exposes a significant vulnerability in UAV-based object detection systems but also lays the groundwork for developing more advanced defense mechanisms.


{
    \bibliographystyle{IEEEtran}
    \bibliography{main}
}


\end{document}